\documentclass[jair,twoside,11pt,theapa]{article}

\usepackage{jair, theapa, rawfonts}

\usepackage{url}
\usepackage{float,color}
\usepackage{epsfig}
\usepackage{epstopdf}
\usepackage{graphicx,subfigure}
\usepackage{threeparttable}
\usepackage{booktabs}
\usepackage{enumitem}
\usepackage{amsmath}
\usepackage{amsfonts}
\usepackage{algorithm}
\usepackage{algorithmic}

\usepackage{helvet}
\usepackage{courier}
\usepackage{amssymb}
\usepackage{multirow}
\usepackage{CJK}
\usepackage{rotating} 

\newcommand\textlcsc[1]{\textsc{\MakeLowercase{#1}}}

\newcommand{\Black}[1]{\textcolor[rgb]{0.00,0.00,0.00}{#1}}
\newcommand{\reviseyq}[1]{\Black{#1}}

\newcommand{\ignore}[1]{}
\usepackage{todonotes}

\ShortHeadings{Cross-lingual Dataless Classification for Languages with Small Wikipedia Presence}
{Song, Mayhew, \& Roth}
\firstpageno{1}

\begin{document}

\title{Cross-lingual Dataless Classification for Languages with Small Wikipedia Presence\thanks{The term ``dataless classification'' was used due to historical reason. To be more concrete, it would be supervisionless, which means that we have no supervised training documents, but still we can do document classification. One would also think about the term labelless, however, we have the label names of the categories.}}

\author{\name Yangqiu Song \email yqsong@cse.ust.hk \\
       \addr Department of Computer Science and Engineering,\\
       Hong Kong University of Science and Technology,\\
       Clear Water Bay, Hong Kong
       \AND
       \name Stephen Mayhew \email mayhew2@illinois.edu \\
       \name Dan Roth \email danr@illinois.edu \\
       \addr Department of Computer Science, \\
       University of Illinois at Urbana-Champaign,\\
       Urbana, IL 21218 USA
}


\maketitle

\begin{abstract}
This paper presents an approach to classify documents in any language into an English topical label space, without any text categorization training data. The approach, Cross-Lingual Dataless Document Classification (CLDDC) relies on mapping the English labels or short category description into a Wikipedia-based semantic representation, and on the use of the target language Wikipedia. Consequently,  performance could suffer when Wikipedia in the target language is small. In this paper, we focus on languages with small Wikipedias, (Small-Wikipedia languages, SWLs). We use a word-level dictionary to convert documents in a SWL to a large-Wikipedia language (LWLs), and then perform CLDDC based on the LWL's Wikipedia. This approach can be applied to {\it thousands of languages}, which can be contrasted with machine translation, which is a supervision heavy approach and can be done for about 100 languages.
We also develop a ranking algorithm that makes use of language similarity metrics to automatically select a good LWL, and show that this significantly improves classification of SWLs' documents, performing comparably to the best bridge possible.
\end{abstract}

\section{Introduction}

Traditional document classification approaches, either monolingual or cross-lingual, use labeled documents to train a model, and apply the model to test documents.
However, when we change the test documents' language or change the label space from one domain to another, we need to re-train the classifiers.
Thus, traditional approaches do not scale well in popular languages, and becomes completely infeasible if we want to classify documents in a small language into any ontology of categories.

Cross-lingual dataless document classification ({\bf CLDDC}) provides a way to classify documents without (re-)training the model, and can thus easily adapt to new domains and many languages~\cite{SongDataless16}.
CLDDC only requires knowing the English name or a short description of each classification category, and works by embedding the label (in English) and the document (in the other language) jointly into the same semantic space.
CLDDC generalizes monolingual dataless classification~\cite{ChangLRS2008,SongR14} by making use of  cross-lingual explicit semantic analysis ({\bf CLESA})~\cite{Potthast2008,Sorg2012}, a generalization of the English explicit semantic analysis ({\bf ESA})~\cite{Gabrilovich2009}.\ignore{It was
	introduced in the context of Information Retrieval in ~\cite{Potthast2008,Sorg2012} and applied to Twitter message classification~\cite{Shirakawa2014}.}
However, CLDDC depends heavily  on the availability of large enough Wikipedia in the documents' language, since it uses the language links in the Wikipedia title space between this language and English. 
Specifically, it has been applied to classify documents in 180 languages that have at least some Wikipedia presence~\cite{SongDataless16}.
There are more than 7,000 known spoken languages, and 2,287 of them have writing systems.\footnote{\url{http://www.alphadictionary.com}}
This means that the CLDDC approach can only work on limited proportion of languages in the world.

\begin{table*}[t]
	\centering
	\small
	\begin{tabular}{llll}
		\toprule
		\textlcsc{Method} & \textlcsc{Resources Available} & \textlcsc{Feasible?} & \textlcsc{Quality}  \\
		\midrule
		CLDDC & Cross-lingual Wiki (179) & Somewhat & So-so \\
		Document Transl. to Eng. & Google Translate (103) + English Wiki (1) & No & Good \\
		Word Transl. to Eng. & Panlex ($>$10,000) + English Wiki (1) & Yes & So-so \\
		Bridged CLDDC & Panlex ($>$10,000) + All Wiki (180) & Yes & Good \\
		\bottomrule
	\end{tabular}
	\caption{ Comparison of different strategies for dataless classification. The numbers in the parentheses are the available numbers of languages. Feasibility means scalable to many languages. Quality means the dataless classification results. Up to date. Panlex has 11,051 language varieties and 6,134 distinctive language codes.}\label{tab:resources}
\end{table*}

This paper tackles the above challenge for languages with little or no presence in Wikipedia, which we call here {\em  small-Wikipedia languages} ({\bf SWLs}).
One can think of multiple ways to facilitate classifying documents in SWLs into an English category ontology. Conceptually, the simplest way is to translate the SWL documents to English or a language with large Wikipedia presence (\textit{large-Wikipedia language}, {\bf LWL}), and then apply English dataless classification or CLDDC for the LWL.
Unfortunately, this requires full document translation which, in turn, requires large amounts of parallel data in the two languages. This is unlikely to be available anytime soon.  
The available resources for training standalone machine translation tools are relatively very sparse.  
For example, Europarl\footnote{\url{http://www.statmt.org/europarl/}} covers 21 languages;
Google Translate\footnote{\url{https://translate.google.com}} covers 103 languages.

The first contribution of this paper is that we show that bi-lingual dictionaries (aka ``word-level translation'')
can be used to support reliable document classification via CLDDC or dataless classification.
This approach scales to many more languages: PanDictionary~\cite{Mausam2010PLT} or later Panlex\footnote{\url{https://panlex.org/}} has word mappings for {\it thousands of languages}.

However, a natural question is, given an SWL document, which bi-lingual dictionary should we use to facilitate good classification into an English category ontology?
Mapping to English may not be the best. 
For example, among the 169 languages in Wiktionary\footnote{\url{https://www.wiktionary.org/}} we can download, there are more than 800 language pairs with more than 1,000 language links, but only 59 of them are associated with English. This analysis, some key figures of it being summarized in Table~\ref{tab:resources}, indicates that in order to facilitate classifying a SWL document into an English ontology, we may need to go through a {\em bridge} language, a LWL for which bi-lingual dictionaries are available.

The second contribution of this paper, is that we show how to choose the best LWL to serve as the bridge language between a given SWL and English.
For example, for Hausa, a SWL, it turns out that if we can find a proper LWL, such as Arabic, then we can use the Arabic-English Wikipedia to perform CLDDC.
Since Arabic is more similar to Hausa compared to English to Hausa, mapping of words from Hausa to Arabic can be better than English.

While the idea of using a bridge (pivot) language is not new~\cite{PaulFS13},
in this paper we: (1) Systematically evaluate CLDDC  using 88 languages, including 39 SWLs and 49 LWLs, and show that this approach successfully supports good classification of a large proportion of SWLs we tested. (2) We propose an automatic way to rank LWLs based on their ability to support good categorization of SWL documents. Specifically, we show how to use RankSVM~\cite{Herbrich2000,Chapelle2010} to learn from the language features to identify which LWLs should be effective as a bridge to a given SWL.
Experiments show that this learning based method is significantly better than the use of handcrafted language similarities to rank the LWLs, and that, in many cases, it selects the best possible bridge.
%

\section{Multilingual 20-newsgroups Data}\label{sec:data}

Since the existing benchmark data sets for multilingual document classification~\cite{Lewis04,hermann2014} focus on a small set of languages, to test the CLDDC in many languages, \reviseyq{we use the data developed by~\cite{SongDataless16}.\footnote{\reviseyq{\url{https://cogcomp.cs.illinois.edu/page/resource_view/104}}}
	They selected 100 documents from 20-newsgroups~\cite{Lang95ICML} which can be 100\% correctly classified using the English dataless classification~\cite{SongR14},
	and used Google Translate API\footnote{\url{https://github.com/mouuff/Google-Translate-API}} to
	translate these documents into 87 languages.\footnote{\reviseyq{We also filtered out Serbo-Croatian (sh) language since it has been deprecated and became a macrolanguage for Croatian (hr), Serbian (sr), Bosnian (bs) and Montenegrin (sr).} }}
We use the English label descriptions for the 20-newsgroups as in~\cite{SongR14}.
Thus, we fixed the English label space with 20 label descriptions.

\begin{figure}[t]
	\centering
	\subfigure[\small Dataless classification on docs. translated back to English. Mean: 0.893; Std: 0.019.]{\label{fig:TranslateBack}
		\includegraphics[width=0.46\textwidth]{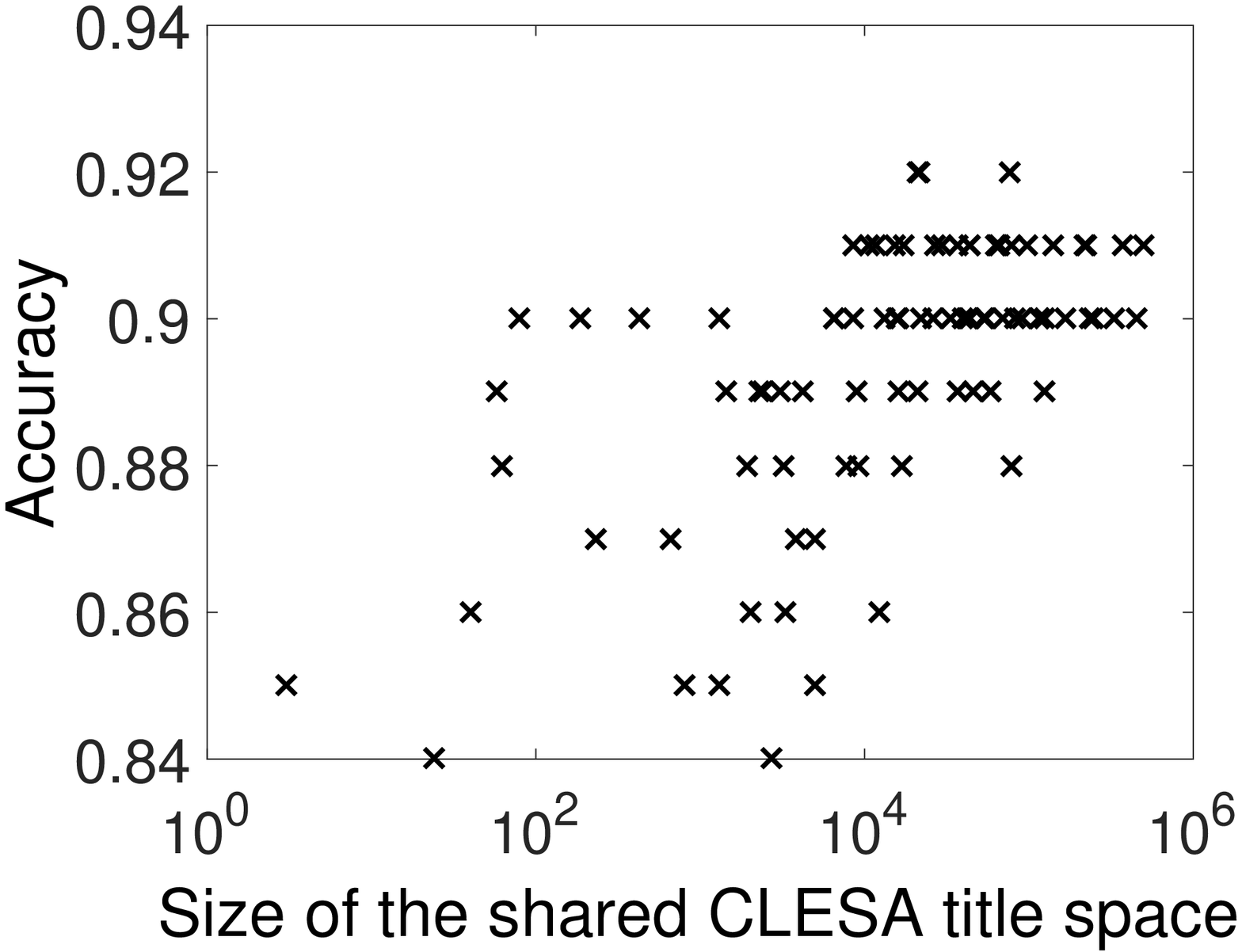}
	} 
	\subfigure[\small CLDDC.]{\label{fig:CLDDC}
		\includegraphics[width=0.46\textwidth]{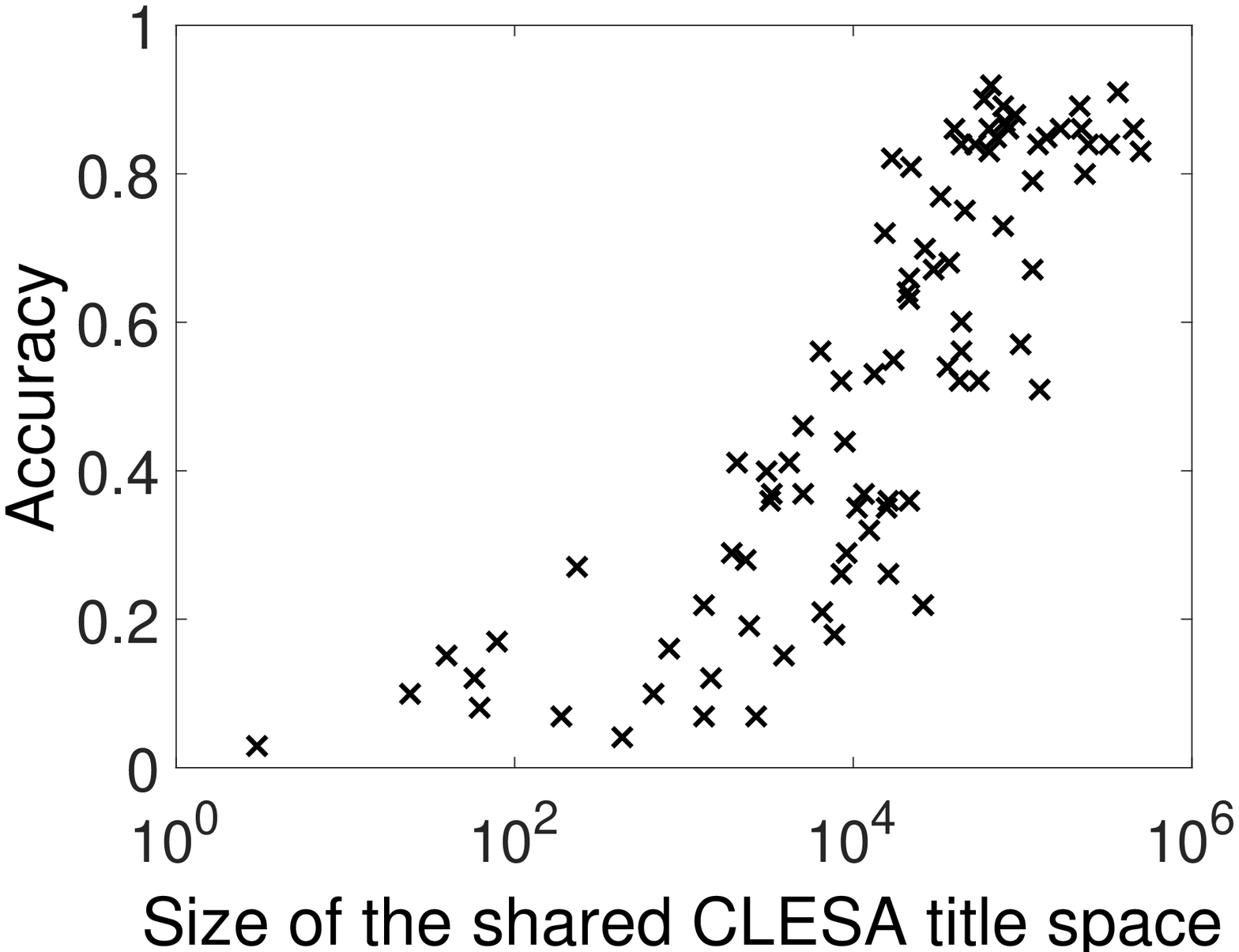}
	}
	\caption{\small Correlation of Wikipedia page log-number and classification accuracy. Each cross represents a language.  (a) $\rho=0.604$, $p=5.6 \times 10^{-10}$. (b) $\rho=0.834$, $p=1.1 \times 10^{-23}$.
		($\rho$: Pearson's correlation coefficient. $p$: the significance value at level 0.05. )}	\label{fig:correlation}
\end{figure}

Suppose that the target documents are in a foreign language $L$ out of the 87 languages.
We use the intersection of the Wikipedia title pages linked between English and $L$ to perform CLESA for CLDDC.
We show the correlation between number of Wikipedia titles used in CLESA and the accuracy of CLDDC for 87 languages in Figure~\ref{fig:CLDDC}.
As shown in Figure~\ref{fig:CLDDC}, the correlation score between the logarithm number of Wikipedia titles used in CLESA and the accuracy for 87 languages is $\rho=0.834$ ($p=1.1 \times 10^{-23}$).
The classification result is significantly correlated with the logarithm number of Wikipedia titles.

To test whether Google translation will hurt the document quality, we perform the following evaluation.
For the 100 documents in each language translated by Google, we translated them back to English again using Google Translate.
Then we performed the English ESA based dataless classification~\cite{SongR14}.
A perfect translation should result in 100\% accuracy.
As shown in Figure~\ref{fig:TranslateBack},
The average classification accuracy is $0.893\pm0.019$, which seems good enough for us to use the translated documents as our evaluation data.
The correlation score between the logarithm number of Wikipedia titles used in CLESA and the accuracy of translated English documents for 87 languages is $\rho=0.604$ ($p=5.6 \times 10^{-10}$).
It seems that Google translate's performance is also correlated with the size of acquired resource.

CLDDC can achieve relatively good performance when the number of Wikipedia titles is large. 
Since the correlation between CLDDC results and Wikipedia sizes is significant, we split the 87 languages based on the classification results.
There are 39 languages with lower than 0.5 classification accuracy, while 48 with higher than 0.5 accuracy.
Figure~\ref{fig:comparison-best} has summarized the best bridged CLDDC for the 39 selected languages.
In this paper, we call the 39 languages the SWLs and the 48 languages as well as English (in total 49) the LWLs.

\section{Bridged Cross-lingual Dataless Classification }

In this section, we introduce the general idea of bridging SWL with LWLs, ans show some comparison results with other unsupervised learning or cross-lingual classification approaches using two typical languages.

\subsection{Bridging SWLs with LWLs}
We first select two typical SWLs, i.e., Hausa and Uzbek, as examples to demonstrate how to use bridging languages to improve dataless classification.
Hausa is a language under the Afro-Asiatic family and further under Chad.
Uzbek is a language under the Middle Turkic family.
Both of the writing systems are related to Arabic and Latin.
The small number of Wikipedia pages, and therefore small shared semantic space, for these two languages (62 for Hausa and 3,082 for Uzbek after intersection with English Wikipedia), means that CLDDC will not be accurate.
\reviseyq{The results are summarized in Table~\ref{tab:hausa}.}
Indeed, classification results are not satisfactory (0.08 for Hausa and 0.40 for Uzbek).

The basic idea of bridged CLDDC is that if we can leverage some word level translation from SWLs to another language, we can use the other language to build ESA/CLESA and further perform dataless classification.
Here we tried to use both English (3 million titles) and Arabic (77,631 intersected titles) to bridge Hausa and Uzbek.
To compare dictionaries, we first used Google Translate to translate all the words (word by word) used in 20-newsgroups documents in Hausa and Uzbek to English.
Then the dataless classification result using English ESA is 0.27 and 0.63 for Hausa and Uzbek respectively.

\begin{table}[t]
	\centering
	{
		\begin{tabular}{lll}
			\toprule
			\textlcsc{Dataless Classif.} & \textlcsc{lg=ha (ACC.)} & \textlcsc{lg=uz (ACC.)}\\
			\midrule
			lg-en Wiki. & 0.08 & 0.40 \\
			en Wiki. (lg-en dict.) & 0.27  & 0.63 \\
			ar-en Wiki. (lg-en dict.) & 0.43 & 0.71 \\
			ar-en Wiki. (lg-ar dict.)  & 0.75 & 0.79 \\
			\midrule
			\textlcsc{One-shot CLSCL} & \textlcsc{lg=ha (ACC.)} & \textlcsc{lg=uz (ACC.)} \\
			\midrule
			en$\rightarrow$lg & 0.088$\pm$0.047 & 0.101$\pm$0.028 \\
			ar$\rightarrow$lg & 0.098$\pm$0.016 & 0.098$\pm$0.038 \\
			\toprule
			\textlcsc{Clustering} & \textlcsc{lg=ha (Pur.)} & \textlcsc{lg=uz (Pur.)} \\
			\midrule
			K-means (en) & 0.714$\pm$0.025 & --\\
			K-means (ar) & 0.698$\pm$0.046 & --\\
			K-means (lg) & 0.684$\pm$0.025 & 0.686$\pm$0.031 \\
			\bottomrule
		\end{tabular}
	}
	\caption{ Comparison on Hausa language data. ``en'' stands for English. ``ar'' stands for Arabic. ``ha'' stands for Hausa. ``uz'' stands for Uzbek. ``ACC.'' stands for accuracy. ``Pur.'' stands for purity.}\label{tab:hausa}
\end{table}

To test the CLDDC using Arabic as a bridging language, we use Google Translate to translate Hausa/Uzbek words into Arabic words.
Then we map each document in Hausa/Uzbek to Arabic, and perform CLESA based on Arabic-English Wikipedia.
The result of dataless classification is 0.75 and 0.79 for Hausa and Uzbek respectively.
We presume that there are two potential reasons for Arabic being better than English.
First, the Arabic-English intersected  space may be less ambiguous than the original English space.
The language links used by CLESA reduce the size of space of Wikipedia titles, but help to disambiguate the semantic meanings.
Second, the word-to-word mappings for Hausa/Uzbek-Arabic are better than those for Hausa/Uzbek-English because Hausa/Uzbek and Arabic are in the same writing system.
To test the two above hypotheses, we also map Hausa/Uzbek to English and use the English part in the Arabic-English intersected Wikipedia to perform dataless classification.
The results are 0.43 and 0.71 respectively, less than using the Arabic part but greater than using Hausa/Uzbek-English mapping for English Wikipedia.
We summarize all the above results in Table~\ref{tab:hausa}.

Moreover, to demonstrate the ability of bridged CLDDC, we also compared with Cross-Language text classification using Structural Correspondence Learning (CLSCL) algorithm~\cite{Prettenhofer2010}, which also uses word level mapping for cross-lingual document classification.
We test on a one-shot learning setting~\cite{FeiFeiFP06,Brenden2015}, which means that only one labeled document per class is used for training.
Dataless classification can also be seen as a one-shot learning but the only example used for each label is just the label name or short description.
To test the algorithm, we split the 100 documents into five folds.
Then we used one fold in source language (e.g., English or Arabic) as supervised data which contains one document per class, used the other four folds in both languages as unsupervised data, and used the other four folds in target language (e.g., Hausa) for testing.
The parameters were set to default values ($\phi=30, m=450$) as in the code\footnote{\url{https://github.com/pprett/nut}} except for the reduced dimensionality $k=3$ which corresponds to the least number of pivot~\cite{Prettenhofer2010} generated by the algorithm in our data.
The mean and std based on five fold cross-validation are also shown in Table~\ref{tab:hausa}.
It seems supervised learning based on only one document per class does not provide reasonably good results, simply
because the known words in the training set cannot cover the words in the test set.
Where as, our CLDDC mapped the original text to a common semantic space represented by Wikipedia, which relatively better overlaps between documents and labels.
\reviseyq{The results in~\cite{SongDataless16} also showed that CLDDC for LWLs is in general comparable to supervised classification trained based on 100 labeled document per class.}

Moreover, we also compare CLDDC with the traditional unsupervised clustering algorithm, K-means, over the TF-IDF features of documents (IDF was computed based each language's 100 documents).
Note that in dataless classification, we only need label names to classify the documents, but K-means needs to know a set of documents in the target language.
\reviseyq{When seeing more documents, CLDDC can also be further improved by bootstrapping~\cite{SongDataless16}.}
We performed ten trials and average the results, using the purity metric to evaluate the accuracy of clustering.
Purity is an average accuracy of each cluster assigned to the max corresponding ground truth label.
It can be regarded as an upper-bound accuracy when we do not know the correspondence between the ground truth labels and the clustered labels.
The clustering results are comparable for English, Arabic, and Hausa, but not as good as bridged CLDDC.
In addition, from the results we can see that there is no clear clue about which language will have better clustering results.

\begin{figure*}[t]
	\centering
	\subfigure[Bridged CLDDC with Google word translation.]{\label{fig:comparison-best}
		\includegraphics[width=0.45\textwidth]{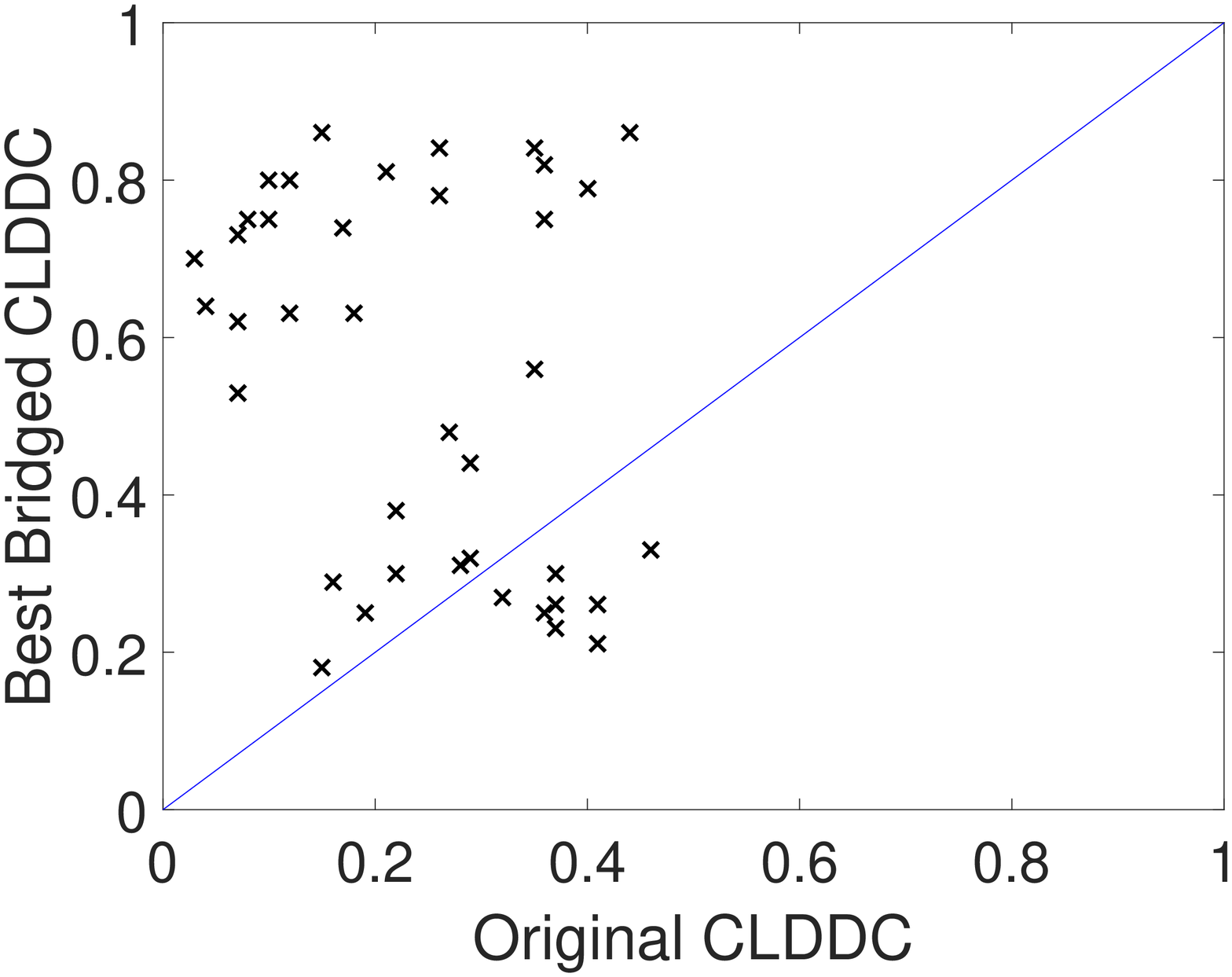}
	}
	\subfigure[Bridged CLDDC with Panlex word translation.]{\label{fig:comparison-panlex-best}
		\includegraphics[width=0.45\textwidth]{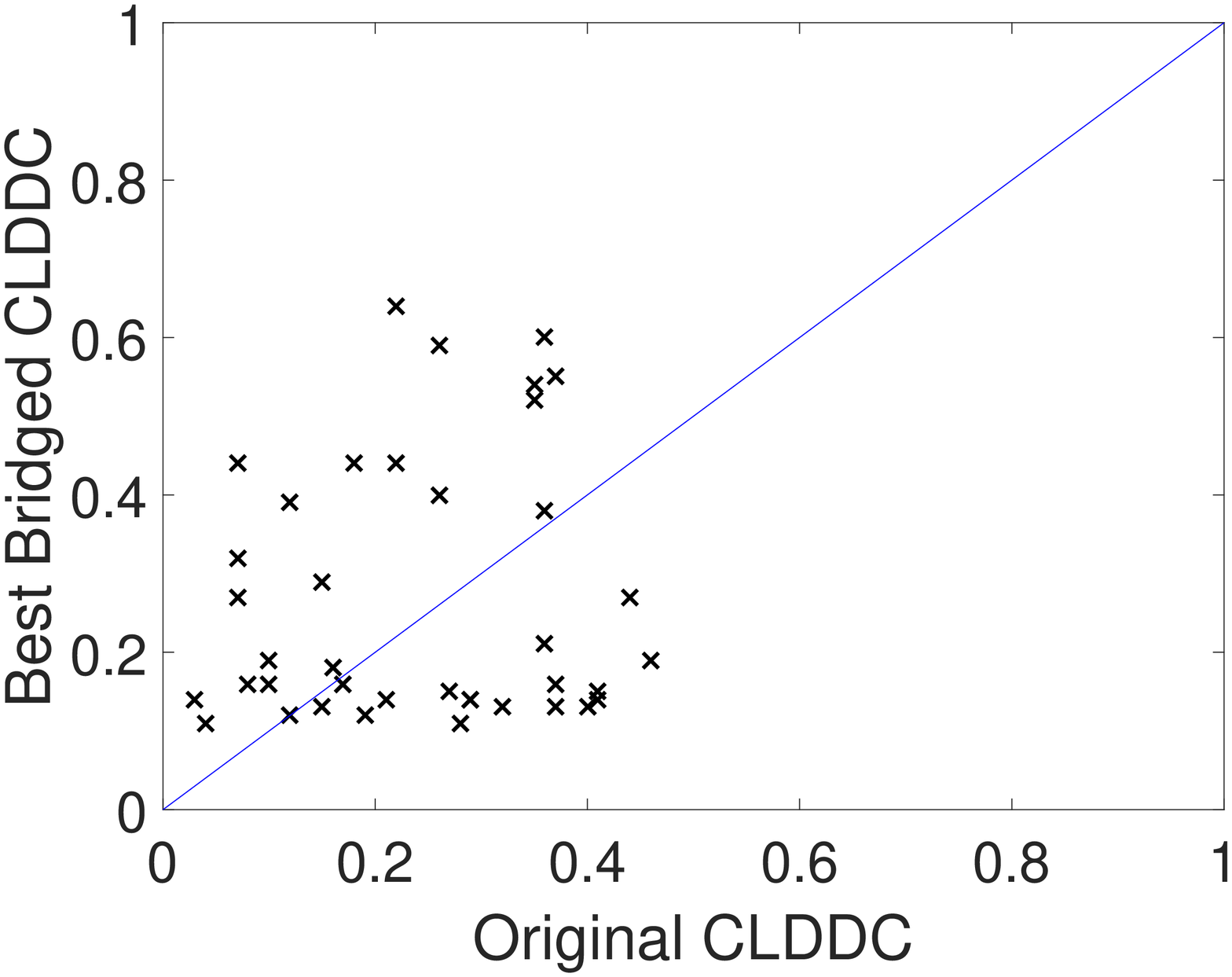}
	}
	\caption{ Comparison of original CLDDC and best bridged CLDDC on 39 SWLs.}	\label{fig:best}
\end{figure*}

\section{Rank Bridging Languages}

Given the fact that for both Hausa and Uzbek, Arabic outperforms English for bridged CLDDC,
and the fact that there are more local languages out of 7,000 languages in the world that cannot be translated to English but may be able to be translated to local popular language,
we want to evaluate which language can be the best language as a bridge for the SWLs.
In Table~\ref{tab:bridging}, we show the top ten bridging languages for the target languages Hausa and Uzbek.
All the translation of words are performed by Google Translate.\footnote{\reviseyq{Here we use Google Translate since it has consistent coverage across the evaluation data we used. Wiktionary (and other dictionaries) covers 1,000+ languages and makes our method a lot more scalable, but we have yet to systematically compare the quality of various dictionaries.}}
The results show that Arabic is the best bridge for both languages.
For all the 39 SWLs, we also checked the bridged results based on all the 49 LWLs, and we selected the best bridging languages and report the classification results.

\begin{figure*}[t]
	\centering
	\includegraphics[width=0.7\textwidth]{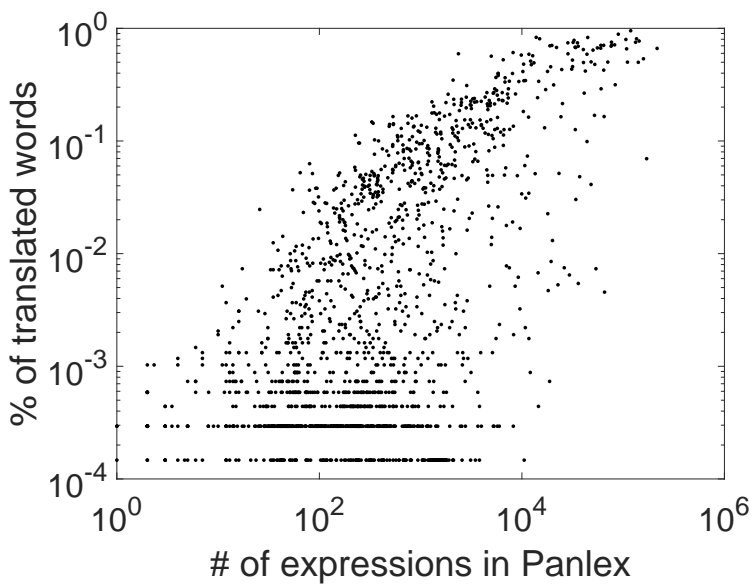}
	\caption{ Panlex word translation on 20-newsgroups data (1,671 languages to English).}	\label{fig:panlex-translation}
\end{figure*}

We compare the original CLDDC results based on the SWL-English Wikpedia with the best LWL bridged CLDDC in Figure~\ref{fig:best}.
We show the results using Google Translate in Figure~\ref{fig:comparison-best}.
We also show the results using Panlex word translation in Figure~\ref{fig:comparison-panlex-best}.
There are 8 out of 39 languages bridged CLDDC being worse than the original CLDDC using Google Translate, while there are 17 languages worse with Panlex translation.
To further evaluate the quality of Panlex dictionaries, we traversed all the 6,134 distinctive language codes in Panlex.
We found there are 1,671 languages with at least one word in the selected 100 documents in 20-newsgroups data that can be translated into English.
The percentage of words that can be translated versus the number of expressions shown in Panlex is shown in Figure~\ref{fig:panlex-translation}.
It turns out that only 12.39\% out of 1,671 languages has more than 10\% words identified.
This is why Panlex translation results are worse than Goolge Translate shown in Figure~\ref{fig:best}.

\begin{table}[t]
	\centering
	\begin{tabular}{lllll}
		\toprule
		&\multicolumn{2}{c}{Hausa} & \multicolumn{2}{c}{Uzbek} \\
		\cmidrule(r){2-3}
		\cmidrule(r){4-5}
		\textsc{Rk.} & \textsc{Bridge} & \textsc{Acc.} & \textsc{Bridge} & \textsc{Acc.} \\
		\midrule
		1  & Arabic & 0.75 & Arabic & 0.79 \\
		2  & Hebrew & 0.72 & Korean & 0.78 \\
		3  & Korean & 0.72 & Hebrew & 0.77 \\
		4  & Bulgarian & 0.67 & Catalan & 0.76 \\
		5  & Persian & 0.67 & Bulgarian & 0.76 \\
		6  & Russian & 0.67 & Russian & 0.71 \\
		7  & Indonesian & 0.65 & Indonesian & 0.70 \\
		8  & Thai & 0.65 & Persian & 0.70 \\
		9  & Japanese & 0.63 & Spanish & 0.70 \\
		10 & Lithuanian & 0.61 & Japanese & 0.70 \\
		\bottomrule
	\end{tabular}
	\caption{ Bridging languages ranks for Hausa and Uzbek translated 20-newsgroups data.}
	\label{tab:bridging}
\end{table}

By analyzing the results in Table~\ref{tab:bridging}, we find some related languages in Latin writing system as Hausa and Uzbek, such as Spanish, Catalan, Indonesian, and Bulgarian, are ranked high.
Some of the top bridging LWLs are in the same family with the target languages.
For example, Arabic, Hebrew, and Hausa are in the Afro-Asiatic family.
Moreover, region of the native speakers is also reflected.
Persian speakers in Iran live relatively near to Uzbek speakers in Uzbekistan.
However, writing system, language family, and region are not the only factors that affects the ranking.
For example, Korean and Japanese are also ranked in the top ten, but none of the above factors appears.
Besides other linguistic factors, we also presume that either the size of Wikipedia or the less ambiguity of translation may help them result in relatively good accuracy.
Then the remaining question is how to automatically select a good bridging LWL for the SWL document classification.

\subsection{Similarity Ranking}
To automatically rank the bridging LWLs for SWLs, we first use the World Atlas of Language Structures (WALS)\footnote{\url{http://wals.info/}} data as language features.
The version we downloaded has 2,679 languages with 198 features including phonological, grammatical, and lexical properties.
We removed latitude and longitude features thus resulting in 196 features.
Given the above analysis, we found four features that are very useful compared to the others, which are {\it genus}, {\it family}, {\it macro area}, and {\it country code}.
From these, we manually developed a similarity value for a pair of languages as follows:
\begin{equation}\label{eq:linguistic}
\begin{array}{ll}
S_{l}(L_1, L_2) & =  50\cdot I_{genus}(L_1, L_2)  \\
& + 50\cdot I_{family}(L_1, L_2) \\
& + 50\cdot I_{macro \; area}(L_1, L_2)  \\
& + 50\cdot I_{country \; code}(L_1, L_2)  \\
& + \sum_{i\in \{others\}} I_{i}(L_1, L_2),
\end{array}
\end{equation}
where $I_{i}(L_1, L_2)=1$ means that two languages $L_1$ and $L_2$ share the same feature $i$.
If one of the most important features is identified, we add a large value into the similarity value.
We set the weight to be 50 with the heuristics that the number can be comparable to the number of 196 features.

We also want to incorporate the size of the Wikipedias since it correlated well with the performance of CLDDC.
Therefore, we rank the languages with the size of Wikipedias:
\begin{equation}\label{eq:wikisize}
\begin{array}{ll}
S_{w}(L)  = \# {\rm Wikipedia \; Title \; in \;} L .
\end{array}
\end{equation}
If Wikipedia size is the only factor, we will always use English as the bridging language.
Besides the Wikipedia size, we also use the language links to rank the bridging languages:
\begin{equation}\label{eq:langlinks}
\begin{array}{ll}
S_{ll}(L_1, L_2)  = \# {\rm Language \; links \; from \;} L_{1}  {\rm \; to \;} L_{2}.
\end{array}
\end{equation}

To combine the two ranking factors, since the scales of two similarity/size values are different,
we first convert each similarity/size value to the rank value.
A larger score denotes a more highly ranked language.
We use $W_{l}$ and $W_{w}$ as the weights for each similarity.
For example, German is ranked as second by $S_{w}$, and there are 49 candidate languages, then $W_{w}({\rm German})=(49-1)/49=0.980$.
Then we use the harmonic mean of two weights as the combined rank value:
\begin{equation}\label{eq:combination}
S_h(L_1, L_2)= \frac{2 W_{l}(L_1, L_2) W_{w}(L_2)}{W_{l}(L_1, L_2) + W_{w}(L_2)},
\end{equation}
where we treat $L_1$ as the SWL and $L_2$ as the bridging LWL.
We use the higher value of $S_h$ to select better bridging LWLs.

\subsection{RankSVM}
The above approaches for ranking the bridging LWLs are handcrafted similarities.
We also tried to use machine learning to learn from the features and generalize to other languages.
Suppose we have a language pair $L_i$ and $L_j$.
We can construct a feature vector ${\bf x}_{ij}$ based on the WALS data, where the $r$th feature is:
\begin{equation}
{\bf x}_{ij}^{(r)} = I_{r}(L_i, L_j),
\end{equation}
where the indicator function denotes both languages sharing the same WALS feature value.

If we consider $L_i$ as the SWL, and there are two candidates LWLs $L_j$ and $L_k$, we can compare $L_j$ and $L_k$ based on their feature vectors by projecting them to a real value: ${\bf w}^T {\bf x}_{ij}$ and ${\bf w}^T {\bf x}_{ik}$.
Thus, if we have a lot of such pairs, we can build a support vector machine to learn the projection vector ${\bf w}$:
\begin{equation}
\frac{1}{2} \|{\bf w}\|^2 + C \sum_{i\in \{SWL\}, j,k\in \{LWL\}} \ell ({\bf w}^T {\bf x}_{ij} - {\bf w}^T {\bf x}_{ik})
\end{equation}
where $\ell$ is the loss function $\ell(t) = \max(0, 1-t)$~\cite{Chapelle2010} and $C$ is the penalty parameter.
Then given the learnt ${\bf w}$, for any pair of LWLs $L_j$ and $L_k$, we can evaluate which one is better to be used to bridge the SWL $L_i$ based on ${\bf w}^T {\bf x}_{ij} - {\bf w}^T {\bf x}_{ik}$.

\begin{sidewaystable}[p]
	\centering
	\begin{tabular}{lllll}
		\toprule
		\textlcsc{Method}         & \textlcsc{Mean$\pm$Std}    & \textlcsc{t-test $p$-value} & \textlcsc{Corr. w. Best (Google)} & \textlcsc{Dependent Corr. $p$-value} \\ \midrule
		Original       & 0.242$\pm$0.126 & -                   & -0.310 & - \\
		Majority Voting& 0.438$\pm$0.236 & $2.061\times 10^{-4}$ & 0.974 &  \\ \midrule
		Best Bridge (Panlex)    & 0.269$\pm$0.167 & 0.406 & -  & - \\
		Best Bridge (Google), upper bound    & 0.546$\pm$0.238 & $2.272\times 10^{-7}$ & 1.000  & - \\ \midrule
		Linguistic     & 0.380$\pm$0.243 & 0.011 & 0.827  & - \\
		Wikipedia language links & 0.275$\pm$0.186 & 0.389 & 0.805  & 0.734  (Ling.) \\
		Wikipedia size & 0.277$\pm$0.186 & 0.366 & 0.856  & 0.607  (Ling.) \\
		Combination (wiki size)    & 0.353$\pm$0.221 & 0.013 & 0.773  & 0.402 (Ling.), 0.096 (Wiki.)  \\
		RankSVM        & 0.465$\pm$0.229 & $2.067\times 10^{-5}$& 0.963  &  $3.646\times 10^{-5}$ (Ling.) \\ 
		\bottomrule
	\end{tabular}
	\caption{ Comparison of different methods to select LWLs to bridge SWL CLDDC.
		The t-tests are performed between each ranking results with the original CLDDC results.
		\reviseyq{The t-test $p$-value between ``RankSVM'' and ``Majority Voting'' is 0.014.}
		The dependent correlation tests are performed with each value with the previous one(s), i.e., ``Wikipedia size vs. Linguistic,'' ``Wikipedia language links vs. Linguistic,'' ``Combination vs. Linguistic (Ling.),'' ``Combination vs. Wikipedia size (Wiki.),'' and ``RankSVM vs. Combination (Comb.).'' All the $p$-values are at 0.05 level (greater than 0.05 will reject the hypothesis).}
	\label{tab:ranking}
\end{sidewaystable}

\subsection{Ranking Results}

Table~\ref{tab:ranking} shows the results.
``Original'' row is the result for original CLDDC.
We compute the mean and standard deviation for the 39 SWLs as well as the correlation of CLDDC with the best bridged CLDDC.
The correlation value is negative.
According to Figure~\ref{fig:comparison-best}, it seems the improvement over smaller original CLDDC accuracies is larger than the ones with larger CLDDC accuracies.

\reviseyq{``Majority Voting'' row shows the results of using all the LWLs to vote for each dataless classification result.
	It shows that ``Majority Voting'' is significantly better than original CLDDC, and highly correlated with ``Best Bridge'' shown in next row.}

``Best Bridge (Google)'' row shows the results of bridged CLDDC results with the best bridge LWLs.
This is the upper-bound of all the other ranking based methods.
We can also see from Figure~\ref{fig:comparison-best} that the result is significantly better.
Although the variance of the results is large, the t-test result still shows significance.
``Best Bridge (Panlex)'' shows no significant improvement over original CLDDC.
However, Panlex has much more languages than Wikipedia and Google Translate. 
Thus, it might be still useful when there is something than no resource at all.

``Linguistic'' row shows the results of bridged LWLs ranked by $S_h(L_1, L_2)$ in Eq.~(\ref{eq:linguistic}).
It is significantly better than original CLDDC at 0.05 level.

``Wikipedia Wikipedia language links'' row shows the results ranked by $S_{ll}(L)$ in Eq.~(\ref{eq:wikisize}).
$S_{ll}(L)$ is almost the same as $S_w(L)$, since for most of the languages, English has the largest language link number.
``Wikipedia size'' row shows the results of bridged LWLs ranked by $S_w(L)$ in Eq.~(\ref{eq:wikisize}).
$S_w(L)$ will always rank English as the bridge language.
We have two interesting findings from the results.
First, the ranking is not significantly better than original CLDDC, and worse than ``Linguistic.''
This means that, bridging SWLs with English by mapping only words may not be a better solution compared to using cross-lingual Wikipedia, even though the cross-lingual Wikipedia is not good enough.
Second, the correlation value between ``Wikipedia size'' and ``Best Bridge'' is higher than the correlation value between ``Linguistic'' and ``Best Bridge.''
However, the dependent correlation test~\cite{howell2011statistical}\footnote{We used the implementation here: \url{https://github.com/psinger/CorrelationStats}.} shows this improvement is not significant.

``Combination'' row shows the results of bridged LWLs ranked by $S_w(L)$ in Eq.~(\ref{eq:combination}).
The results show that combining the ``Linguistic'' and ``Wikipedia size'' features by hand shows no improvement over pure ``Linguistic'' features.

``RankSVM'' row shows the results using RankSVM.
We split the SWLs into five folds.
Then we perform a five-fold cross validation to generate the results.
For each validation, we use 80\% of the SWLs as training data, where each language has 49 LWLs accuracies.
We use the 49 accuracies to generate $49\times 48 / 2$ pairs.
Then we use the learnt model to rank the other 20\% SWLs.
After the five-fold cross validation, we can rank all the SWLs based on the each learnt model.
We tune the parameter of $C$ using a grid search in $\{10^{-2}, 10^{-1}, \ldots, 10^{4}\}$.
The average result over 39 SWLs is significantly better than original CLDDC  ($p=2.067\times 10^{-5}$) and ``Majority Voting'' ($p=0.014$).
The correlation with ``Best Bridge'' is also significantly better than ``Linguistic'' with ``Best Bridge.''
This means that machine learning based method is significantly better than the unsupervised voting and ranking with handcrafted similarities.
By looking into the averaged weights of RankSVM in five fold cross validation, we select the five top weights with largest absolute values, which are:
``Internally-headed relative clauses'' (-0.3613),
``Front Rounded Vowels'' (0.1656),
``Absence of Common Consonants'' (0.1538),
``Optional Double Negation in SVO languages'' (-0.1402),
``Number of Genders'' (0.1385).

\section{Related Work}
\label{sec:relatedwork}
In this section, we briefly survey some related work.

\subsection{Cross-lingual Classification}
\label{sec:relatedwork-crosslingual}
Cross-lingual document classification has attracted more attention recently in low-resource settings.
where target language training data is minimal or unavailable.
It is a natural sub-topic of transfer learning~\cite{Pan2010}.
In cross-lingual document classification, we train a classifier on labeled documents in the {\it source} language, and classify documents in the {\it target} language.
Existing approaches either need a parallel corpus to train
word embeddings for different languages~\cite{hermann2014},
require labeled documents in both source and target
languages~\cite{XiaoG13},
make use of machine translation techniques to
{translate words~\cite{Prettenhofer2010}} or documents~\cite{AminiG10}, {or combine different approaches~\cite{Shi2010}}.
Among the existing approaches, word translation is the cheapest way, while document translation and annotation on the target domain are the most expensive.
In the middle, parallel or comparable corpora may be used to learn a good word/document representation, which avoids document translation but can still find a correspondence between source and target languages.

The strength of cross-lingual document classification is that it can be generalized to multiple languages even in the absence of resources.
However, when we change the label space from one domain to another, we should perform translation again and re-train the classifiers.
Instead of using parallel corpus to train a classifier or train a translation model, our approach only needs the comparable corpus, Wikipedia, in different languages aligned with English.
Then if a user can tell the name of the category, cross-lingual dataless classification can perform text classification on-the-fly.
For LWLs, CLDDC is comparable to supervised learning method with about 100 to 200 labeled document per label~\cite{SongDataless16}.
CLDDC does not need training data in source language. Instead, it only needs the label names and the comparable corpus, Wikipedia, in different languages aligned with English.
For LWLs, CLDDC is comparable to supervised learning method with about 100 to 200 labeled document per label~\cite{SongDataless16}.

\subsection{Cross-lingual Representation Learning}
Clearly cross-lingual dataless classification is related to representation learning.
Our current approach is based on ESA~\cite{Gabrilovich2009}, 
which is built based on Wikipedia inverted index.
Essentially, ESA is a distributional representation of document-level context of words.
It regards each Wikipedia page as a concept corresponding to an entity, category, or topic.
Then each word is represented by its related concepts.
Cross-lingual ESA leverages the language links in Wikipeida to relate pages in other languages to English.
Then words or texts in two different languages (English and another) can be mapped to the same semantic space represented by Wikipedia concepts.
Recently, motivated by the simplicity and success of neural network based word embedding~
\cite{MikolovYZ13,Mikolov132},
multi-lingual~\cite{AlRfouPS13},
or cross-lingual~\cite{klementiev2012,XiaoG13,hermann2014,FaruquiD14,LuWBGL15,UpadhyayFDR16}
representation learning are also investigated.
Other representations such as cross-lingual topic models~\cite{Mimno2009,PlattTY10,Zhang2010}
or Brown clusters~\cite{Tackstrom2012}
were also studied.
Similar to cross-lingual classification, these representation learning approaches  need either parallel corpora~\cite{klementiev2012,hermann2014}, some labeled data in the target domain~\cite{XiaoG13}, or words being (partially) aligned in a dictionary
\cite{Zhang2010}.

It has been shown that CLESA outperforms one of popular the cross-lingual embedding approach~\cite{hermann2014} on two benchmark datasets for CLDDC~\cite{SongDataless16}.

\subsection{Pivot based Machine Translation}
Pivot language is used to help machine translation when there is no enough resources to train a translation model from source language to target language~\cite{MannY01,CohnL07,UtiyamaI07,WuW09,LeuschMCN10,PaulFS13}.
For example, Paul et al.,~\cite{PaulFS13} used 22 Indo-European and Asian languages to evaluate how to select a good pivot language for machine translation. They evaluated 45 features falling into eight categories. Besides the language family feature, they used more translation-relevant features such as length of sentence, reordering, overlap of vocabulary, etc. They showed that the final result is mostly affected by the source-pivot and pivot-target translation performance. They also mentioned machine learning based method in the future work, but we are unaware of a follow-up paper that succeeded in doing it.

Different from machine translation which needs the sentence level source-pivot and pivot-target translation, in cross-lingual classification, it is sufficient to use word dictionaries, making borrowing a bridging language more scalable to many languages and thus more practically useful.
To the best of our knowledge, we have studied largest number of LWLs (49) and SWLs (39) with largest number of linguistic features (196).

\subsection{Zero/One-shot Learning}
Zero-shot learning
~\cite{PalatucciPHM09,SocherGMN13,Elhoseiny13,RomeraParedesT15}
and one-shot learning~\cite{FeiFeiFP06,Brenden2015}
were
first introduced in the computer vision community and are now recognized by
the natural language processing
community~\cite{YazdaniH15,LazaridouDB15}.
One-shot learning requires one example for training, while
in zero-shot learning,
the test data is different from the training data (e.g., a new label
space).
However, in contrast to the dataless scenario, both learning protocols
require some training data. The dataless classification protocol, on
the other hand, assumes no direct training data but only the label names or descriptions.
Compared to one-shot learning, the labels can be relatively simpler.
In addition, it relies on background
data from external knowledge sources (like Wikipedia), that is used in
an unsupervised way to generate a common semantic space.

\section{Conclusion}
We studied the problem of CLDDC for SWLs.
CLDDC uses English labels to classify documents in other languages and is scalable to many languages and adaptive to any label space.
However, if Wikipedia for a language is not large enough, the performance is not acceptable.
In this paper, we simply map the words in SWL documents to LWL words, and perform dataless classification based on LWLs.
We systematically evaluate 39 SWLs and 49 LWLs.
Experiments show that bridging the SWLs with LWLs can significantly improve the classification results.
Moreover, learning from the existing ranking results can be generalized to other languages.

\acks{
	This work was supported by DARPA under agreement numbers HR0011-15-2-0025 and FA8750-13-2-0008. The U.S. Government is authorized to reproduce and distribute reprints for Governmental purposes notwithstanding any copyright notation thereon.  The views and conclusions contained herein are those of the authors and should not be interpreted as necessarily representing the official policies or endorsements, either expressed or implied, of any of the organizations that supported the work.
}


\begin{thebibliography}{}
	
	\bibitem[\protect\BCAY{Al-Rfou', Perozzi,\ \BBA\ Skiena}{Al-Rfou'
		et~al.}{2013}]{AlRfouPS13}
	Al-Rfou', R., Perozzi, B., \BBA\ Skiena, S. \BBOP2013\BBCP.
	\newblock \BBOQ Polyglot: Distributed word representations for multilingual
	nlp\BBCQ\
	\newblock In {\Bem CoNLL}, \BPGS\ 183--192.
	
	\bibitem[\protect\BCAY{Amini\ \BBA\ Goutte}{Amini\ \BBA\
		Goutte}{2010}]{AminiG10}
	Amini, M.\BBACOMMA\  \BBA\ Goutte, C. \BBOP2010\BBCP.
	\newblock \BBOQ A co-classification approach to learning from multilingual
	corpora\BBCQ\
	\newblock {\Bem Machine Learning}, {\Bem 79\/}(1-2), 105--121.
	
	\bibitem[\protect\BCAY{Chang, Ratinov, Roth,\ \BBA\ Srikumar}{Chang
		et~al.}{2008}]{ChangLRS2008}
	Chang, M.-W., Ratinov, L., Roth, D., \BBA\ Srikumar, V. \BBOP2008\BBCP.
	\newblock \BBOQ Importance of semantic representation: Dataless
	classification\BBCQ\
	\newblock In {\Bem AAAI}, \BPGS\ 830--835.
	
	\bibitem[\protect\BCAY{Chapelle\ \BBA\ Keerthi}{Chapelle\ \BBA\
		Keerthi}{2010}]{Chapelle2010}
	Chapelle, O.\BBACOMMA\  \BBA\ Keerthi, S.~S. \BBOP2010\BBCP.
	\newblock \BBOQ Efficient algorithms for ranking with {SVMs}\BBCQ\
	\newblock {\Bem Inf. Retr.}, {\Bem 13\/}(3), 201--215.
	
	\bibitem[\protect\BCAY{Cohn\ \BBA\ Lapata}{Cohn\ \BBA\ Lapata}{2007}]{CohnL07}
	Cohn, T.\BBACOMMA\  \BBA\ Lapata, M. \BBOP2007\BBCP.
	\newblock \BBOQ Machine translation by triangulation: Making effective use of
	multi-parallel corpora\BBCQ\
	\newblock In {\Bem ACL}.
	
	\bibitem[\protect\BCAY{Elhoseiny, Saleh, ,\ \BBA\ A.Elgammal}{Elhoseiny
		et~al.}{2013}]{Elhoseiny13}
	Elhoseiny, M., Saleh, B., , \BBA\ A.Elgammal \BBOP2013\BBCP.
	\newblock \BBOQ Write a classifier: Zero shot learning using purely textual
	descriptions\BBCQ\
	\newblock In {\Bem ICCV}, \BPGS\ 1433--1441.
	
	\bibitem[\protect\BCAY{Faruqui\ \BBA\ Dyer}{Faruqui\ \BBA\
		Dyer}{2014}]{FaruquiD14}
	Faruqui, M.\BBACOMMA\  \BBA\ Dyer, C. \BBOP2014\BBCP.
	\newblock \BBOQ Improving vector space word representations using multilingual
	correlation\BBCQ\
	\newblock In {\Bem EACL}, \BPGS\ 462--471.
	
	\bibitem[\protect\BCAY{Gabrilovich\ \BBA\ Markovitch}{Gabrilovich\ \BBA\
		Markovitch}{2009}]{Gabrilovich2009}
	Gabrilovich, E.\BBACOMMA\  \BBA\ Markovitch, S. \BBOP2009\BBCP.
	\newblock \BBOQ Wikipedia-based semantic interpretation for natural language
	processing\BBCQ\
	\newblock {\Bem Journal of Artificial Intelligence Research}, {\Bem 34\/}(1),
	443--498.
	
	\bibitem[\protect\BCAY{Herbrich, Graepel,\ \BBA\ Obermayer}{Herbrich
		et~al.}{2000}]{Herbrich2000}
	Herbrich, R., Graepel, T., \BBA\ Obermayer, K. \BBOP2000\BBCP.
	\newblock {\Bem Large margin rank boundaries for ordinal regression}.
	\newblock MIT Press, Cambridge, MA.
	
	\bibitem[\protect\BCAY{Hermann\ \BBA\ Blunsom}{Hermann\ \BBA\
		Blunsom}{2014}]{hermann2014}
	Hermann, K.~M.\BBACOMMA\  \BBA\ Blunsom, P. \BBOP2014\BBCP.
	\newblock \BBOQ Multilingual models for compositional distributed
	semantics\BBCQ\
	\newblock In {\Bem ACL}, \BPGS\ 58--68.
	
	\bibitem[\protect\BCAY{Howell}{Howell}{2011}]{howell2011statistical}
	Howell, D.~C. \BBOP2011\BBCP.
	\newblock {\Bem Statistical methods for psychology}.
	\newblock Cengage Learning.
	
	\bibitem[\protect\BCAY{Klementiev, Titov,\ \BBA\ Bhattarai}{Klementiev
		et~al.}{2012}]{klementiev2012}
	Klementiev, A., Titov, I., \BBA\ Bhattarai, B. \BBOP2012\BBCP.
	\newblock \BBOQ Inducing crosslingual distributed representations of
	words\BBCQ\
	\newblock In {\Bem COLING}, \BPGS\ 1459--1474.
	
	\bibitem[\protect\BCAY{Lake, Salakhutdinov,\ \BBA\ Tenenbaum}{Lake
		et~al.}{2015}]{Brenden2015}
	Lake, B.~M., Salakhutdinov, R., \BBA\ Tenenbaum, J.~B. \BBOP2015\BBCP.
	\newblock \BBOQ Human-level concept learning through probabilistic program
	induction\BBCQ\
	\newblock {\Bem Science}, {\Bem 350\/}(6266), 1332--1338.
	
	\bibitem[\protect\BCAY{Lang}{Lang}{1995}]{Lang95ICML}
	Lang, K. \BBOP1995\BBCP.
	\newblock \BBOQ Newsweeder: Learning to filter netnews\BBCQ\
	\newblock In {\Bem ICML}, \BPGS\ 331--339.
	
	\bibitem[\protect\BCAY{Lazaridou, Dinu,\ \BBA\ Baroni}{Lazaridou
		et~al.}{2015}]{LazaridouDB15}
	Lazaridou, A., Dinu, G., \BBA\ Baroni, M. \BBOP2015\BBCP.
	\newblock \BBOQ Hubness and pollution: Delving into cross-space mapping for
	zero-shot learning\BBCQ\
	\newblock In {\Bem ACL}, \BPGS\ 270--280.
	
	\bibitem[\protect\BCAY{Leusch, Max, Crego,\ \BBA\ Ney}{Leusch
		et~al.}{2010}]{LeuschMCN10}
	Leusch, G., Max, A., Crego, J.~M., \BBA\ Ney, H. \BBOP2010\BBCP.
	\newblock \BBOQ Multi-pivot translation by system combination\BBCQ\
	\newblock In {\Bem 2010 International Workshop on Spoken Language Translation,
		{IWSLT} 2010, Paris, France, December 2-3, 2010}, \BPGS\ 299--306.
	
	\bibitem[\protect\BCAY{Lewis, Yang, Rose,\ \BBA\ Li}{Lewis
		et~al.}{2004}]{Lewis04}
	Lewis, D.~D., Yang, Y., Rose, T.~G., \BBA\ Li, F. \BBOP2004\BBCP.
	\newblock \BBOQ Rcv1: A new benchmark collection for text categorization
	research\BBCQ\
	\newblock {\Bem J. Mach. Learn. Res.}, {\Bem 5}, 361--397.
	
	\bibitem[\protect\BCAY{Li, Fergus,\ \BBA\ Perona}{Li et~al.}{2006}]{FeiFeiFP06}
	Li, F., Fergus, R., \BBA\ Perona, P. \BBOP2006\BBCP.
	\newblock \BBOQ One-shot learning of object categories\BBCQ\
	\newblock {\Bem {IEEE} Trans. Pattern Anal. Mach. Intell.}, {\Bem 28\/}(4),
	594--611.
	
	\bibitem[\protect\BCAY{Lu, Wang, Bansal, Gimpel,\ \BBA\ Livescu}{Lu
		et~al.}{2015}]{LuWBGL15}
	Lu, A., Wang, W., Bansal, M., Gimpel, K., \BBA\ Livescu, K. \BBOP2015\BBCP.
	\newblock \BBOQ Deep multilingual correlation for improved word
	embeddings\BBCQ\
	\newblock In {\Bem {NAACL}-{HLT}}, \BPGS\ 250--256.
	
	\bibitem[\protect\BCAY{Mann\ \BBA\ Yarowsky}{Mann\ \BBA\
		Yarowsky}{2001}]{MannY01}
	Mann, G.~S.\BBACOMMA\  \BBA\ Yarowsky, D. \BBOP2001\BBCP.
	\newblock \BBOQ Multipath translation lexicon induction via bridge
	languages\BBCQ\
	\newblock In {\Bem {NAACL}}.
	
	\bibitem[\protect\BCAY{Mausam, Soderland, Etzioni, Weld, Reiter, Skinner,
		Sammer,\ \BBA\ Bilmes}{Mausam et~al.}{2010}]{Mausam2010PLT}
	Mausam, Soderland, S., Etzioni, O., Weld, D.~S., Reiter, K., Skinner, M.,
	Sammer, M., \BBA\ Bilmes, J. \BBOP2010\BBCP.
	\newblock \BBOQ Panlingual lexical translation via probabilistic
	inference\BBCQ\
	\newblock {\Bem Artif. Intell.}, {\Bem 174\/}(9-10), 619--637.
	
	\bibitem[\protect\BCAY{Mikolov, Sutskever, Chen, Corrado,\ \BBA\ Dean}{Mikolov
		et~al.}{2013a}]{Mikolov132}
	Mikolov, T., Sutskever, I., Chen, K., Corrado, G.~S., \BBA\ Dean, J.
	\BBOP2013a\BBCP.
	\newblock \BBOQ Distributed representations of words and phrases and their
	compositionality\BBCQ\
	\newblock In {\Bem NIPS}, \BPGS\ 3111--3119.
	
	\bibitem[\protect\BCAY{Mikolov, Yih,\ \BBA\ Zweig}{Mikolov
		et~al.}{2013b}]{MikolovYZ13}
	Mikolov, T., Yih, W.-t., \BBA\ Zweig, G. \BBOP2013b\BBCP.
	\newblock \BBOQ Linguistic regularities in continuous space word
	representations.\BBCQ\
	\newblock In {\Bem HLT-NAACL}, \BPGS\ 746--751.
	
	\bibitem[\protect\BCAY{Mimno, Wallach, Naradowsky, Smith,\ \BBA\
		McCallum}{Mimno et~al.}{2009}]{Mimno2009}
	Mimno, D., Wallach, H.~M., Naradowsky, J., Smith, D.~A., \BBA\ McCallum, A.
	\BBOP2009\BBCP.
	\newblock \BBOQ Polylingual topic models\BBCQ\
	\newblock In {\Bem EMNLP}, \BPGS\ 880--889.
	
	\bibitem[\protect\BCAY{Palatucci, Pomerleau, Hinton,\ \BBA\ Mitchell}{Palatucci
		et~al.}{2009}]{PalatucciPHM09}
	Palatucci, M., Pomerleau, D., Hinton, G.~E., \BBA\ Mitchell, T.~M.
	\BBOP2009\BBCP.
	\newblock \BBOQ Zero-shot learning with semantic output codes\BBCQ\
	\newblock In {\Bem NIPS}, \BPGS\ 1410--1418.
	
	\bibitem[\protect\BCAY{Pan\ \BBA\ Yang}{Pan\ \BBA\ Yang}{2010}]{Pan2010}
	Pan, S.~J.\BBACOMMA\  \BBA\ Yang, Q. \BBOP2010\BBCP.
	\newblock \BBOQ A survey on transfer learning\BBCQ\
	\newblock {\Bem IEEE Trans. on Knowledge and Data Engineering}, {\Bem
		22\/}(10), 1345--1359.
	
	\bibitem[\protect\BCAY{Paul, Finch,\ \BBA\ Sumita}{Paul
		et~al.}{2013}]{PaulFS13}
	Paul, M., Finch, A.~M., \BBA\ Sumita, E. \BBOP2013\BBCP.
	\newblock \BBOQ How to choose the best pivot language for automatic translation
	of low-resource languages\BBCQ\
	\newblock {\Bem {ACM} Trans. Asian Lang. Inf. Process.}, {\Bem 12\/}(4), 14.
	
	\bibitem[\protect\BCAY{Platt, Toutanova,\ \BBA\ tau Yih}{Platt
		et~al.}{2010}]{PlattTY10}
	Platt, J.~C., Toutanova, K., \BBA\ tau Yih, W. \BBOP2010\BBCP.
	\newblock \BBOQ Translingual document representations from discriminative
	projections\BBCQ\
	\newblock In {\Bem EMNLP}, \BPGS\ 251--261.
	
	\bibitem[\protect\BCAY{Potthast, Stein,\ \BBA\ Anderka}{Potthast
		et~al.}{2008}]{Potthast2008}
	Potthast, M., Stein, B., \BBA\ Anderka, M. \BBOP2008\BBCP.
	\newblock \BBOQ A wikipedia-based multilingual retrieval model\BBCQ\
	\newblock In {\Bem ECIR}, \BPGS\ 522--530.
	
	\bibitem[\protect\BCAY{Prettenhofer\ \BBA\ Stein}{Prettenhofer\ \BBA\
		Stein}{2010}]{Prettenhofer2010}
	Prettenhofer, P.\BBACOMMA\  \BBA\ Stein, B. \BBOP2010\BBCP.
	\newblock \BBOQ Cross-language text classification using structural
	correspondence learning\BBCQ\
	\newblock In {\Bem ACL}, \BPGS\ 1118--1127.
	
	\bibitem[\protect\BCAY{Romera{-}Paredes\ \BBA\ Torr}{Romera{-}Paredes\ \BBA\
		Torr}{2015}]{RomeraParedesT15}
	Romera{-}Paredes, B.\BBACOMMA\  \BBA\ Torr, P. H.~S. \BBOP2015\BBCP.
	\newblock \BBOQ An embarrassingly simple approach to zero-shot learning\BBCQ\
	\newblock In {\Bem ICML}, \BPGS\ 2152--2161.
	
	\bibitem[\protect\BCAY{Shi, Mihalcea,\ \BBA\ Tian}{Shi et~al.}{2010}]{Shi2010}
	Shi, L., Mihalcea, R., \BBA\ Tian, M. \BBOP2010\BBCP.
	\newblock \BBOQ Cross language text classification by model translation and
	semi-supervised learning\BBCQ\
	\newblock In {\Bem EMNLP}, \BPGS\ 1057--1067.
	
	\bibitem[\protect\BCAY{Socher, Ganjoo, Manning,\ \BBA\ Ng}{Socher
		et~al.}{2013}]{SocherGMN13}
	Socher, R., Ganjoo, M., Manning, C.~D., \BBA\ Ng, A.~Y. \BBOP2013\BBCP.
	\newblock \BBOQ Zero-shot learning through cross-modal transfer\BBCQ\
	\newblock In {\Bem NIPS}, \BPGS\ 935--943.
	
	\bibitem[\protect\BCAY{Song\ \BBA\ Roth}{Song\ \BBA\ Roth}{2014}]{SongR14}
	Song, Y.\BBACOMMA\  \BBA\ Roth, D. \BBOP2014\BBCP.
	\newblock \BBOQ On dataless hierarchical text classification\BBCQ\
	\newblock In {\Bem AAAI}, \BPGS\ 1579--1585.
	
	\bibitem[\protect\BCAY{Song, Upadhyay, Peng,\ \BBA\ Roth}{Song
		et~al.}{2016}]{SongDataless16}
	Song, Y., Upadhyay, S., Peng, H., \BBA\ Roth, D. \BBOP2016\BBCP.
	\newblock \BBOQ Cross-lingual dataless classification for many languages\BBCQ\
	\newblock In {\Bem IJCAI}, \BPGS\ 2901--2907.
	
	\bibitem[\protect\BCAY{Sorg\ \BBA\ Cimiano}{Sorg\ \BBA\
		Cimiano}{2012}]{Sorg2012}
	Sorg, P.\BBACOMMA\  \BBA\ Cimiano, P. \BBOP2012\BBCP.
	\newblock \BBOQ Exploiting wikipedia for cross-lingual and multilingual
	information retrieval\BBCQ\
	\newblock {\Bem Data and Knowledge Engineering}, {\Bem 74}, 26--45.
	
	\bibitem[\protect\BCAY{T\"{a}ckstr\"{o}m, McDonald,\ \BBA\
		Uszkoreit}{T\"{a}ckstr\"{o}m et~al.}{2012}]{Tackstrom2012}
	T\"{a}ckstr\"{o}m, O., McDonald, R., \BBA\ Uszkoreit, J. \BBOP2012\BBCP.
	\newblock \BBOQ Cross-lingual word clusters for direct transfer of linguistic
	structure\BBCQ\
	\newblock In {\Bem NAACL-HLT}, \BPGS\ 477--487.
	
	\bibitem[\protect\BCAY{Upadhyay, Faruqui, Dyer,\ \BBA\ Roth}{Upadhyay
		et~al.}{2016}]{UpadhyayFDR16}
	Upadhyay, S., Faruqui, M., Dyer, C., \BBA\ Roth, D. \BBOP2016\BBCP.
	\newblock \BBOQ Cross-lingual models of word embeddings: An empirical
	comparison\BBCQ\
	\newblock {\Bem ACL}.
	
	\bibitem[\protect\BCAY{Utiyama\ \BBA\ Isahara}{Utiyama\ \BBA\
		Isahara}{2007}]{UtiyamaI07}
	Utiyama, M.\BBACOMMA\  \BBA\ Isahara, H. \BBOP2007\BBCP.
	\newblock \BBOQ A comparison of pivot methods for phrase-based statistical
	machine translation\BBCQ\
	\newblock In {\Bem NAACL-HLT}, \BPGS\ 484--491.
	
	\bibitem[\protect\BCAY{Wu\ \BBA\ Wang}{Wu\ \BBA\ Wang}{2009}]{WuW09}
	Wu, H.\BBACOMMA\  \BBA\ Wang, H. \BBOP2009\BBCP.
	\newblock \BBOQ Revisiting pivot language approach for machine
	translation\BBCQ\
	\newblock In {\Bem ACL/IJCNLP}, \BPGS\ 154--162.
	
	\bibitem[\protect\BCAY{Xiao\ \BBA\ Guo}{Xiao\ \BBA\ Guo}{2013}]{XiaoG13}
	Xiao, M.\BBACOMMA\  \BBA\ Guo, Y. \BBOP2013\BBCP.
	\newblock \BBOQ Semi-supervised representation learning for cross-lingual text
	classification.\BBCQ\
	\newblock In {\Bem EMNLP}, \BPGS\ 1465--1475.
	
	\bibitem[\protect\BCAY{Yazdani\ \BBA\ Henderson}{Yazdani\ \BBA\
		Henderson}{2015}]{YazdaniH15}
	Yazdani, M.\BBACOMMA\  \BBA\ Henderson, J. \BBOP2015\BBCP.
	\newblock \BBOQ A model of zero-shot learning of spoken language
	understanding\BBCQ\
	\newblock In {\Bem EMNLP}, \BPGS\ 244--249.
	
	\bibitem[\protect\BCAY{Zhang, Mei,\ \BBA\ Zhai}{Zhang et~al.}{2010}]{Zhang2010}
	Zhang, D., Mei, Q., \BBA\ Zhai, C. \BBOP2010\BBCP.
	\newblock \BBOQ Cross-lingual latent topic extraction\BBCQ\
	\newblock In {\Bem ACL}, \BPGS\ 1128--1137.
	
\end{thebibliography}
\end{document}